\documentclass[conference,letterpaper]{IEEEtran}
\IEEEoverridecommandlockouts
\usepackage{amsmath,amssymb,amsfonts}
\usepackage{algorithmic}
\usepackage{graphicx}
\usepackage{textcomp}
\usepackage{hyperref}
\usepackage[OT1]{fontenc}
\usepackage{pifont}
\usepackage{multirow}
\usepackage{threeparttable}
\usepackage{booktabs}% 
\usepackage{subcaption} % Put this in your preamblehttp://ctan.org/pkg/booktabs
\usepackage{float}

\usepackage[dvipsnames,svgnames]{xcolor, colortbl}

\definecolor{LightCyan}{rgb}{0.88,1,1}
\def\BibTeX{{\rm B\kern-.05em{\sc i\kern-.025em b}\kern-.08em
    T\kern-.1667em\lower.7ex\hbox{E}\kern-.125emX}}
\begin{document}
\bstctlcite{IEEEexample:BSTcontrol}
%%----------------------------------------------------------------------------------%%
%% 							TITLE AND LIST OF AUTHOR'S
%%----------------------------------------------------------------------------------%%
\title{Isolated Sign Language Recognition with Segmentation and Pose Estimation\\
}
%%%---------------- Uncomment for Blind Review -------------------%%
%\author{
%1\textsuperscript{st} Name Surname,
%2\textsuperscript{nd} Name Surname,
%}
%\IEEEauthorblockA{
%\textit{name of organization (of Aff.)}\\
%City, Country \\
%email address}
%}
%%%---------------- Uncomment for Blind Review -------------------%%
%---------------- Uncomment for Final Version -------------------%%
% \author{
% \IEEEauthorblockN{
% Danny Perkins,
% Dhrumil Patel,
% Davis Hunter,
% Galen Flanagan,
% Tanmoy Saha
% }
% \IEEEauthorblockA{
% \textit{Organisation}\\
% City, Country \\
% email}
% }

\author{
\IEEEauthorblockN{Daniel Perkins, Davis Hunter, Dhrumil Patel, and Galen Flanagan}

\IEEEauthorblockA{ University of Tennessee, Knoxville}
}

%---------------- Uncomment for Final Version -------------------%%
\maketitle
%%----------------------------------------------------------------------------------%%
%% 									PAPER ABSTRACT
%%----------------------------------------------------------------------------------%%
\begin{abstract}
The recent surge in large language models has automated translations of spoken and written languages. However, these advances remain largely inaccessible to American Sign Language (ASL) users, whose language relies on complex visual cues. Isolated sign language recognition (ISLR)—the task of classifying videos of individual signs—can help bridge this gap, but is currently limited by scarce per-sign data, high signer variability, and substantial computational costs. We propose a model for ISLR that reduces computational requirements while maintaining robustness to signer variation. Our approach integrates (i) a pose estimation pipeline to extract hand and face joint coordinates, (ii) a segmentation module that isolates relevant information, and (iii) a ResNet–Transformer backbone to jointly model spatial and temporal dependencies.
\end{abstract}

%%----------------------------------------------------------------------------------%%
%% 									PAPER KEYWORDS
%%----------------------------------------------------------------------------------%%
% \begin{IEEEkeywords}
% Keywords
% \end{IEEEkeywords}

%%----------------------------------------------------------------------------------%%
%% 								MAIN CONTENT OF THE PAPER
%%----------------------------------------------------------------------------------%%
\section{Introduction and Problem Statement}
\label{sec:Introduction}

\subsection{Background}

With the advent of modern large language models (LLMs), communication between people from diverse linguistic and cultural backgrounds has become significantly more accessible. Recent advancements, such as GPT‑4o \cite{openai2024gpt4ocard} and Voxtral \cite{liu2025voxtral} mark a major step toward universal translation. However, they are fundamentally limited to textual and audio inputs. With over 500,000 people ASL signers in the U.S. \cite{cdhh_asl_2025}, there is a significant need for automatic sign language translation. Unlike other languages, ASL relies on dynamic visual cues expressed through hand movements and facial expressions. 

This paper focuses on isolated sign language recognition (ISLR), where the goal is to translate videos of individual signs. Given a dataset of ASL signals with clear start and end points, the goal is to classify each video sequence into one word, from a predefined vocabulary.

Although ISLR has made significant progress, it is still not accurate enough to replace human interpreters. Sign language data presents a great challenge, as its input is represented by videos that span both the visual and temporal domains. Furthermore, although current datasets contain tens of thousands of videos, the vast size of the ASL vocabulary means that most individual words are represented only a few dozen times. This limitation restricts the robustness of ISLR models, making it challenging for them to generalize effectively to unique ASL accents, signer ethnicities, and video backgrounds.

\subsection{Problem Statement}

We introduce an efficient deep learning architecture that reduces computational requirements while maintaining robustness to visual and signer variations. The proposed model is the first we know of that integrates all three of the following:
\begin{itemize}
    \item A pose estimation pipeline that extracts joint coordinates from the hands and face,
    \item A segmentation module that masks out irrelevant pixels, isolating the signer’s hands and face,
    \item A hybrid ResNet–Transformer backbone that captures both spatial and temporal dependencies.
\end{itemize}
\section{Related Work}

\subsection{The Dataset}
\label{sec:the_dataset}

We focus our analysis on Microsoft's \href{https://www.microsoft.com/en-us/research/project/asl-citizen/}{ASL Citizen dataset}. This dataset has 83,399 videos with a total of 2,731 unique ASL glosses (vocab words). The videos are crowdsourced, featuring signs from 51 individuals with diverse backgrounds and varying levels of ASL fluency. Roughly 63\% of the video generators are female, and 37\% are male. For a more thorough analysis of the structure of the dataset, refer to Appendix \ref{the_dataset}.

\subsection{Visual and Temporal Models}

ISLR requires a complex model capable of fusing the visual and temporal domains. There has been a significant amount of research in deep learning models for visual and time-series based input. ResNets \cite{he2015deepresiduallearningimage} and Vision Transformers (ViTs) \cite{dosovitskiy2021imageworth16x16words} serve as strong backbones for feature extraction in vision models. Additionally, LSTMs \cite{hochreiter1997long} and Transformers \cite{vaswani2023attentionneed} are widely used for temporal understanding. 

Although these models achieve competitive results in their respective input domains, it is difficult to fuse them into a model capable of understanding both visual and temporal inputs simultaneously. In an attempt to resolve this issue, \cite{tran2015learningspatiotemporalfeatures3d} introduced 3D ConvNets which uses $3\times3\times3$ kernels to capture both spatial and temporal information among the pixels. However, their reliance on fixed-size temporal windows limit their ability to model long-range temporal dependencies. TimeSformers \cite{bertasius2021spacetimeattentionneedvideo} address this issue by leveraging self-attention. But, their substantial demand for computational resources hinders practical deployment for real-time translation.

\subsection{Human Pose Estimation}
\label{Pose_Estimation}

Human Pose Estimation (HPE) is a computer vision task that identifies and tracks the spatial coordinates of key human joints and landmarks from images or videos. Instead of processing dense RGB pixel data, HPE models output a lightweight, structured representation of the body's configuration. For sign language recognition, this typically includes keypoints of the hands and face. This joint data provides a computationally efficient and signer-invariant representation, separating the signer from visual noise like complex backgrounds or varied lighting.

A prominent tool for this task is Google's MediaPipe \cite{mediapipeholistic, bazarevsky2022blazeposeghum}. It employs a multi-stage approach optimized for real-time performance. First, a body pose detector (e.g., BlazePose) identifies the person's location and main body joints. This initial pose estimation is then used to derive regions of interest for the hands and face. By running specialized models on these smaller cropped regions, the pipeline tracks joints in the hands and face. This approach provides a comprehensive set of keypoints for all relevant sign language movements, making it a common preprocessing step in modern ISLR systems.

\subsection{Isolated Sign Language Recognition (ISLR): Current Methods}

Early ISLR approaches use CNNs to extract spatial-temporal features from video frames, and LSTMs or Transformers to model temporal dependencies between frames. More recently, transformer models with self-supervised pre-training—such as Feat/VideoMAE variants—have emerged as the leading platform for RGB video understanding \cite{sandoval2023ssvt, zuo2023nla, woods2023encoderonly, vvitwlasl2025}. These approaches incorporate language-aware objectives, improving performance by effectively capturing long-range dependencies in the data. However, these methods require substantial computational resources and are sensitive to background and lighting deviations.

Other ISLR approaches use human pose estimation to reduce visual variability by representing the signer through joint coordinates. SAM-SLR processes this skeleton data by integrating a GCN with spatiotemporal convolutions and fusing pose features with RGB input \cite{jiang2021sam}. Subsequent transformer-based architectures \cite{bohacek2022signpose} and bidirectional GCNs with graph-attention formulations \cite{dafnis2022bidirectional} have further improved results. Although these pose-based pipelines are more signer-agnostic and efficient, they struggle with occlusion and fail to understand fine-grained texture details.

There has also been work which focuses the visual signals by masking out the background (segmentation). Two practical methods are (i) landmark-driven ROIs that crop hands/face for subsequent classification, \cite{mediapipeholistic, bazarevsky2022blazeposeghum} and (ii) learned localizers, such as the YOLO family, to
detect or segment articulators \cite{tian2025yolov12}. These cropped frames can improve the robustness of ISLR models by reducing their tendencies to overfit to irrelevant details in the background. However, inaccuracies in the segmentation limit the overall accuracy. 

We build upon these approaches by developing a pipeline that simultaneously processing segmented images and joint coordinate data.
\section{Methodological Development}

\subsection{Preprocessing}

\begin{figure}[ht]
    \centering
    \begin{subfigure}[b]{0.49\linewidth}
        \centering
        \includegraphics[width=\linewidth]{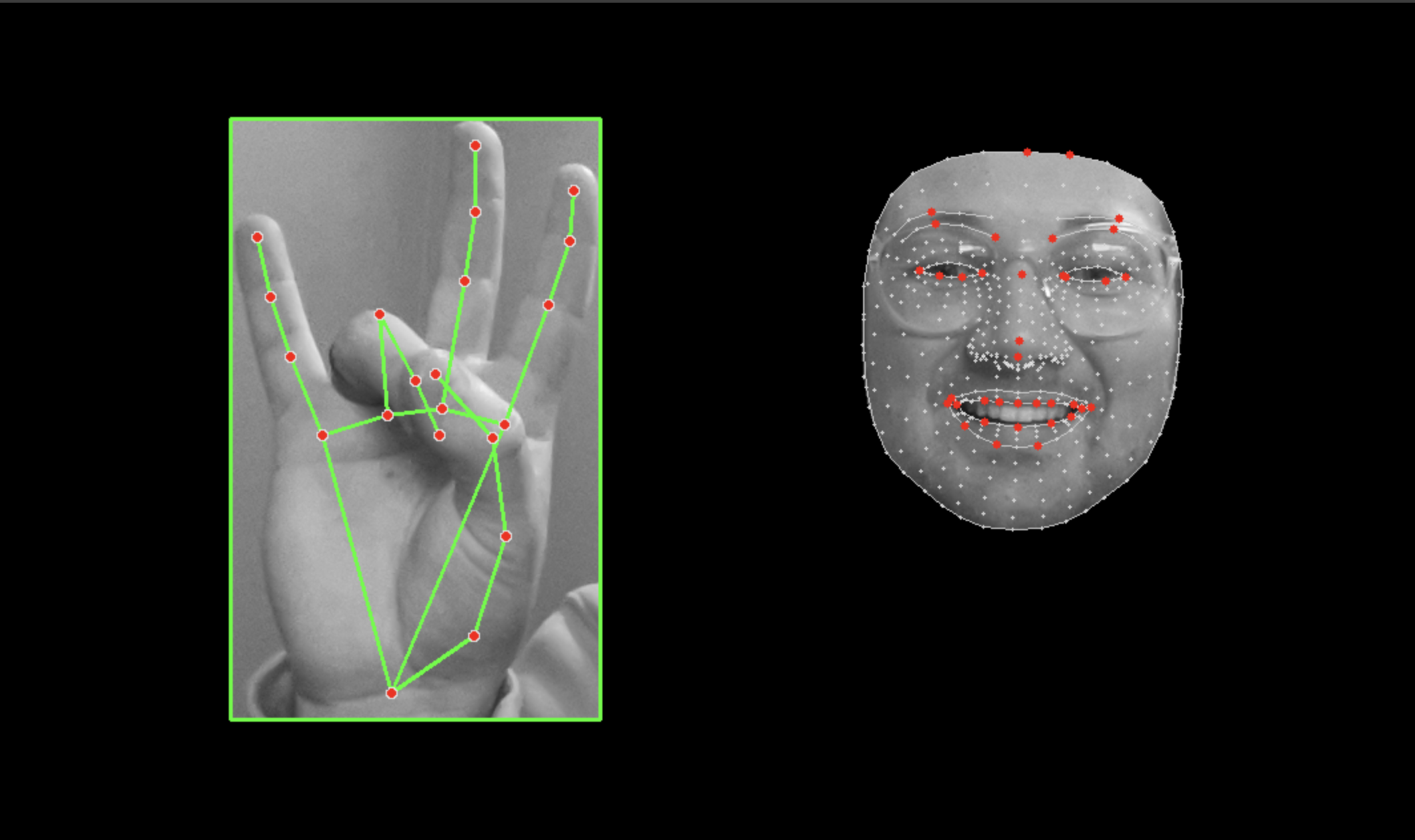}
        % \caption{Subcaption for left image} % Optional: add subcaption text here
    \end{subfigure}
    \hfill
    \begin{subfigure}[b]{0.49\linewidth}
        \centering
        \includegraphics[width=\linewidth]{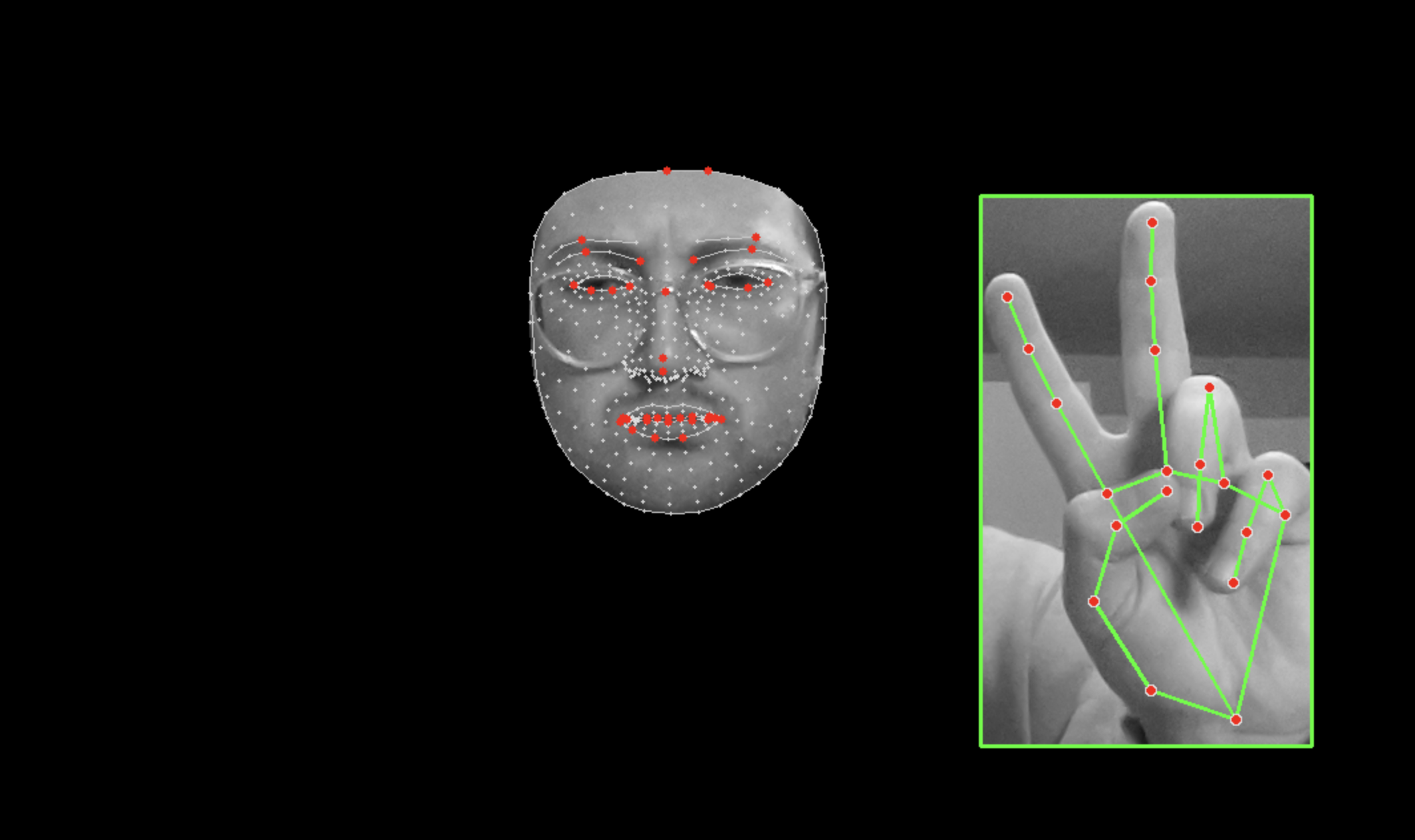}
        % \caption{Subcaption for right image} % Optional: add subcaption text here
    \end{subfigure}
    \caption{Examples of frames after pose processing. The joint coordinates of the hand and face are output and used to crop out the hands and face.}
    \label{fig:preprocessing}
\end{figure}

Since skin and background color are not informative, we first convert the RGB pixels of each frame to grayscale and normalize the values to be between $0$ and $1$. Then, to improve training efficiency and effectiveness, we
extract sign-language-relevant features from each frame by isolating the hands and head and obtaining information about the location of the joints. As explained in Section \ref{Pose_Estimation}, we employ Google's Mediapipe to extract the joint coordinates. Then, as shown in Figure \ref{fig:preprocessing}, those coordinates are used to segment the image by zeroing out all the pixels corresponding to the irrelevant background. 

Through video normalization, pose estimation, and segmentation, input variability from changes in background and signer is effectively removed, allowing the model to generalize more robustly to unseen data. Moreover, reducing the dimensionality of the input data facilitates faster training and inference.

\subsection{Proposed Architecture}
\label{sec:proposed_model}

\begin{figure}[ht]
    \begin{center}
        \includegraphics[width=1\linewidth]{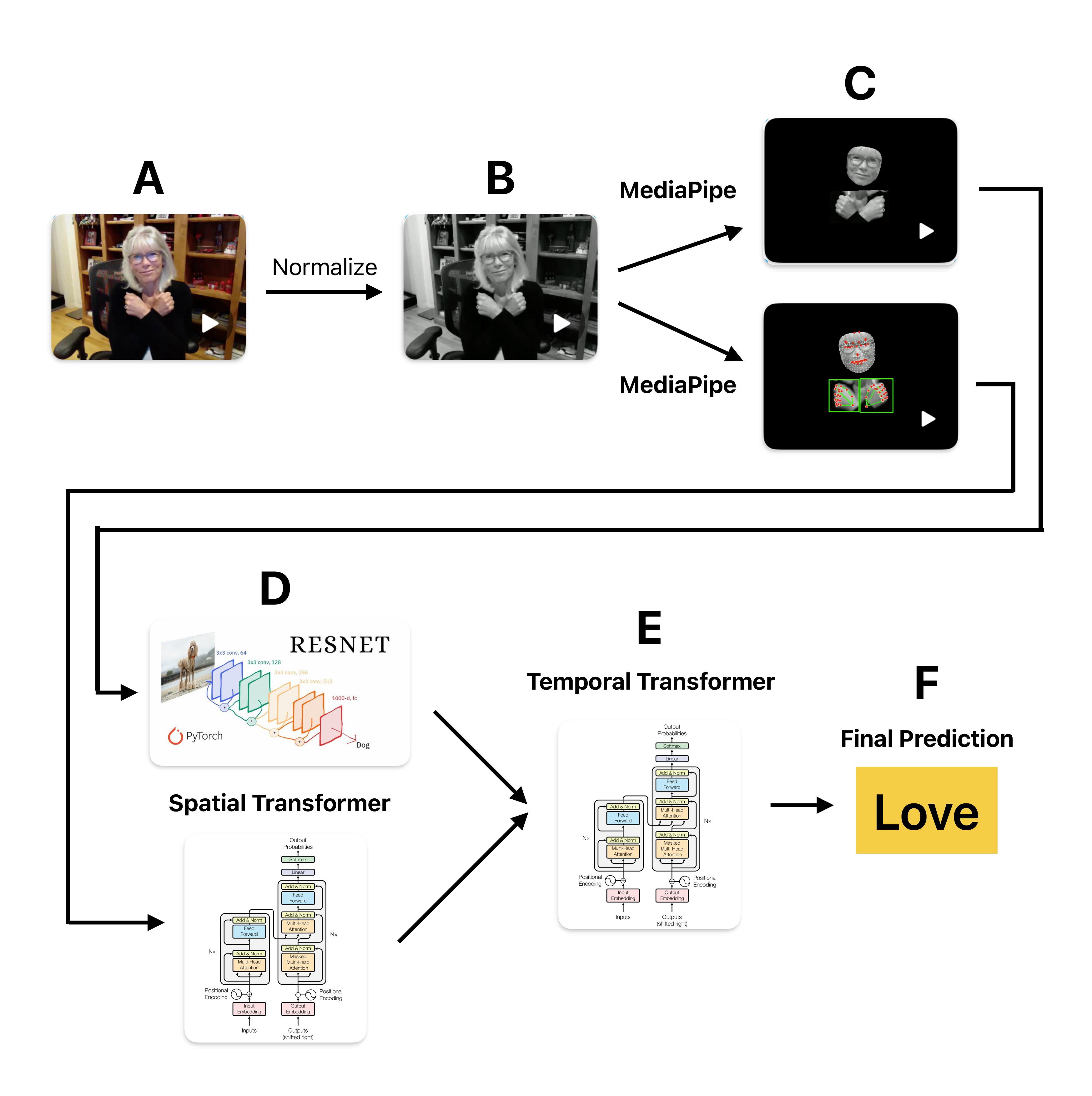}
        \caption{Illustration of the original proposed model; (A) RGB video input; (B) Normalized video (C) Segmented video and joint coordinates from MediaPipe; (D) In each frame, the segmented video is passed into a ResNet and the coordinates are passed into a transformer; (E) The embeddings from each frame are concatenated and passed into a transformer; (F) After a linear layer, the final prediction is made.}\label{fig:the_model}
    \end{center}
    \vspace{-1em}
\end{figure}

The pipeline for the entire model is visualized in Figure \ref{fig:the_model}. After preprocessing, each segmented frame sequence is paired with its corresponding set of normalized joint coordinates. Visual features are extracted from each frame using a ViT or ResNet pre-trained on ImageNet. In parallel, joint coordinates for each frame, structured as arrays of normalized $(x,y,z)$ values, are passed into a transformer that captures spatial dependencies among keypoints, generating spatial embeddings. 

The visual and spatial embeddings are then concatenated across the temporal dimension to form a fused multimodal representation. The resulting sequence of embeddings is processed by an LSTM or Transformer, which models the dynamic evolution of actions across frames. Finally, mean pooling is applied across the sequence to yield a compact representation, which is passed through a linear layer and softmax activation to obtain the final prediction. 

\subsection{Modifications}

To uncover the contributions of individual components and enhance overall predictive accuracy, we carefully refine the model beyond the initial baseline. By investigating the effects of each type of input, normalization, computer vision models, time-series models, and regularization, we provide deeper insight into the strengths and weaknesses of our architecture.

\subsubsection{Video Input}

After normalizing the videos and passing them through MediaPipe, two streams of grayscale video data are generated: the original normalized sequence and a corresponding segmented sequence. Initially, the proposed architecture sought to exploit the pixel information by ingesting both streams as input to 3D Vision Transformers, enabling joint modeling of spatial and temporal dependencies within the video data. However, the computational cost associated with 3D ViTs rendered this approach impractical.

Consequently, the methodology was revised to process video frames individually, generating frame-level embeddings. Since direct training of conventional Vision Transformer (ViT) models was found to be infeasible with our limited computational resources, we employed the ResNet18. Unfortunately, as shown in Section \ref{res:video}, the ResNet exhibited poor generalization performance on the validation set. Since each vocabulary word has only about 15 corresponding videos in the dataset, the high variance of the network induced overfitting. 

Based on these findings, all pixel-based visual inputs were excluded from subsequent experiments. The downstream models rely solely on the pose coordinates produced by MediaPipe, a robust model pre-trained on an extensive visual dataset.

\subsubsection{Normalization}\label{mod:normalization}

Raw coordinate data from MediaPipe is highly sensitive to the subject's position within the frame and their distance from the camera. To ensure the model learns the sign's motion rather than the signer's location, we experimented with three distinct normalization strategies to achieve translation and scale invariance.

First, we implemented a \textit{Global Nose-Anchored Normalization}. This method calculated the position of the nose tip for each frame and treated it as the origin $(0,0,0)$ for the entire video sequence. While this preserved global motion, it proved brittle. Jitter in the nose detection caused the entire skeleton to shake artificially, introducing noise.

Second, we attempted a \textit{Face-Centered Normalization}. To mitigate the jitter of a single keypoint, we calculated the center of mass of all 37 face landmarks. While statistically more stable, using a single global anchor point for the duration of the video occasionally failed to account for significant shifts in the user's stance over time.

Finally, we implemented \textit{Per-Frame Center of Mass Normalization}. In this approach, we calculate the center of mass of all visible joints (hands and face) for every individual frame and subtract this vector from the coordinates. Additionally, we scale the coordinates of each frame such that the maximum deviation falls within the range $[-1, 1]$. Empirically, this method yielded the highest performance. By re-centering every frame, the model was forced to focus exclusively on the relative configuration of the limbs and hand shapes, removing all noise related to body sway or camera movement.

\subsubsection{LSTM}

After the joint coordinates are normalized, the next step is to pass the features through a temporal model. We first pass the coordinates of each frame through a linear network, to obtain more refined features. Then, those features are passed into an LSTM, a model designed to account for both long-term and short-term temporal dependencies. This generates embeddings for each frame which can then be pooled together via mean pooling to represent the entire sequence as a single embedding. We implement both a standard LSTM, which processes input in the forward direction only, and a bidirectional LSTM, which processes input in both forward and backward directions.  

\subsubsection{Temporal Transformer}

In addition to the LSTM, we explored the efficacy of the Transformer architecture applied to pose estimation, often referred to as a PoseTransformer. Unlike LSTMs, which process data sequentially, Transformers utilize a self-attention mechanism that allows the model to weigh the importance of different frames in the sequence simultaneously, regardless of their distance in time. Theoretically, this allows the model to better capture long-range dependencies—such as the relationship between the starting hand shape and the final resting position of a sign—without suffering from the vanishing gradient problem often seen in RNNs.

We implemented a PoseTransformer with positional encodings, a model dimension of 256, 4 attention heads, and 3 encoder layers. In our experiments, the PoseTransformer demonstrated strong generalization capabilities, achieving a validation accuracy of approximately $60\%$. The results for the model with various architectures for the temporal component are explained in Section \ref{temporal_model_results}.

\subsubsection{Dropout}
\label{mod:dropout}

In order to further prevent overfitting, we utilize dropout \cite{srivastava2014dropout}. Dropout randomly deactivates a fraction of nodes in each layer during training to prevent overreliance on any single feature or connection. By temporarily zeroing out certain neurons, the network is encouraged to distribute feature learning across multiple units, reducing the likelihood of overfitting. Dropout increases the bias of deep neural networks while reducing their variance. Therefore, selecting an appropriate rate is critical to achieve an optimal bias-variance tradeoff. In Section \ref{res:dropout}, we train the model with various dropout proportions to empirically determine the optimal rate.

\subsubsection{Stochastic Data Augmentation}
\label{mod:augmentation}

Given the limited size of the dataset, preventing the model from memorizing specific training examples was a primary challenge. To address this, we implemented a stochastic augmentation pipeline. We applied random transformations to the normalized pose data, including \textit{jittering} (adding Gaussian noise to coordinates), \textit{scaling} (randomly resizing the skeleton by a factor of 0.8 to 1.2), and \textit{temporal dropout} (randomly zeroing out a small percentage of frames to simulate occlusion or tracking failure).
% TODO: Add figures to appendix
% \input{Current_Progress_And_Preliminary_Results/Current_Progress_And_Preliminary_Results}
% \input{Next_Steps_And_Project_Plan/Next_Steps_And_Project_Plan}
% Next two sections only needed for final report
\section{Complete Results}

We evaluate our model on the validation set using accuracy and top 5 accuracy, which measures the percent of datapoints that had the correct label in one of the top 5 model predictions. 

\subsection{Original Results}
\label{org_results}

As shown in Appendix \ref{app_original_model}, the original model failed to surpass baseline accuracy. Furthermore, it required a substantial amount of training time. Even when deployed on the HPC, available RAM was insufficient to load the entire corpus of 80,000 videos into memory. Consequently, the data loader had to read video data from storage at each training iteration, leading to significant I/O bottlenecks.

Additionally, the dataset's large vocabulary size, comprising 2,731 unique glosses, each represented by approximately 15 instances, intensified the challenges of effective learning. Under these circumstances, the neural network exhibited catastrophic forgetting, where knowledge acquired from earlier presentations of individual glosses was overwritten before completing an epoch \cite{Kirkpatrick_2017}. This necessitates prolonged training over hundreds epochs to stabilize learning, which simply was not feasible with our limited computational resources.

\subsection{Downsampling}

As a result of our limited computational resources, we were compelled to downsample the dataset. This facilitated rapid testing of our model while still allowing us to demonstrate its effectiveness. To develop a deployable model for real-world applications, future research should focus on training our model on the entire dataset using distributed GPU resources.

In order to downsample the data we found the 100 most common glosses (words) in the training dataset. In cases of ties we used alphabetical order to decide most common gloss. In addition, we skipped any words that were already in the list under a different version. This resulted in a training dataset with 1800 videos, a validation data set with 365 videos, and a test set with 1286 videos. This significant reduction in the size of the dataset alleviated the issue of catastrophic forgetting, which in turn reduced both the number of epochs required for training and the duration of each epoch. Therefore, training could be done on local computers with limited GPU resources.

\subsection{Proposed Model}

\subsubsection{Video Input}
\label{res:video}

As shown in Appendix \ref{app_proposed_model}, the original model (Figure \ref{fig:the_model}) trained on the downsampled dataset reaches a plateau after about 15 epochs, with training and validation accuracies of 1.4\% and 1.18\%, respectively. These results only marginally exceed the 1\% baseline, suggesting that the segmented video input was too noisy for the complex ResNet18 model to learn meaningful representations. The poor performance also suggests the possibility of an underlying issue in the video pre-processing or training pipeline for the ResNet architecture, motivating us to evaluate the model using coordinate data alone

\subsubsection{Normalization}

Table \ref{tab:normalization_results} illustrates the impact of coordinate preprocessing on model performance. While the raw feature vectors contain the necessary spatial information, they are heavily encumbered by variance in the signer's position within the frame and their distance from the camera. As shown in the results, effectively isolating the sign's relative motion through \textit{Per-Frame Center of Mass} normalization resulted in a validation accuracy increase of approximately 16\% over the raw baseline. In contrast, less stable methods, such as anchoring to a single oscillating joint (the nose), actively hindered learning, demonstrating that robust normalization is a prerequisite for convergence in skeleton-based ISLR.

\begin{table}[H]
    \centering
    \begin{tabular}{llc}
    \hline
    \textbf{Model} & \textbf{Normalization Strategy} & \textbf{Val Acc} \\
    \hline
    Bi-LSTM & None (Raw Coordinates) & 50.29\% \\
    Bi-LSTM & Global Nose-Anchored & 39.0\% \\
    Bi-LSTM & Global Head-Anchored & 47.98\% \\
    Bi-LSTM & \textbf{Per-Frame Center of Mass} & \textbf{66.18\%} \\
    \hline
    \end{tabular}
    \caption{Impact of Normalization Strategies on Validation Accuracy. Using raw data resulted in moderate performance, while unstable normalization (Global Nose-Anchored) actively harmed model convergence. The Per-Frame Center of Mass approach yielded the best results for both architectures.}
    \label{tab:normalization_results}
\end{table}

\subsubsection{Temporal Model}
\label{temporal_model_results}

Table \ref{results:temporal} displays the results for various implementations of the temporal model. As expected, the bidirectional LSTM (with normalized joint coordinates) had the highest accuracy and top 5 accuracy. Thus, for ISLR, it is important to account for both the forward and reverse direction in the temporal domain.

\begin{table}[H]
    \centering
    \begin{tabular}{lcccc}
    \hline
    \textbf{Model} & \textbf{Bidirectional} & \textbf{Normalize} & \textbf{Acc} & \textbf{Top5 Acc} \\
    \hline
    LSTM & No & No & 45.66\% & 78.90\%\\
    LSTM & Yes & No & 50.29\% &79.19\% \\
    LSTM (Trial 1) & Yes & Yes &61.27\% & 86.99\% \\
    LSTM (Trial 2) & Yes & Yes &65.61\% & 89.01\% \\
    \textbf{LSTM (Trial 3)} & \textbf{Yes} & \textbf{Yes} & \textbf{66.18\%} & \textbf{89.60\%} \\
    PoseTransformer & N/A & Yes & 60.40\% & 85.67\% \\
    \hline \\
\end{tabular}
\caption{Validation results on the downsampled dataset for the model with various temporal architectures (with and without normalization) and a dropout of $0.2$. The bidirectional LSTM is the best model, achieving an accuracy as high as 66.18\%. The PoseTransformer with dropout of $0.1$ and normalization achieved a validation accuracy of $60.4\%$ competing with the LSTM mode. With more testing and fine tuning, the PoseTransformer may outperform the LSTM.}
\label{results:temporal}
\end{table}

\subsubsection{Dropout}
\label{res:dropout}

Table \ref{results:dropout} displays the results for the model trained with various percentages of dropout (Section \ref{mod:dropout}). Generalization was highest when 10\% of the nodes in each layer were randomly dropped out.

\begin{table}[H]
    \centering
    \begin{tabular}{lccc}
    \hline
    \textbf{Dropout} & \textbf{Accuracy} & \textbf{Top5 Accuracy} \\
    \hline
    0\% & 61.56\% & 85.26\%\\
    \textbf{10\%} & \textbf{68.50\%} &\textbf{89.60\%} \\
    20\% & 66.47\% &89.60\% \\
    30\% & 63.87\% &87.28\% \\
    
\hline \\
    \end{tabular}
\caption{Validation results on the downsampled dataset for the model with various values for dropout. Each model only takes in the pose coordinates, normalizes those coordinates. and has a bidirectional LSTM for the temporal dimension.}
\label{results:dropout}
\end{table}

\subsubsection{Final Model}

\begin{figure}[h]
    \centering
    \begin{subfigure}[t]{1.0\linewidth}
        \centering
        \includegraphics[width=1\linewidth]{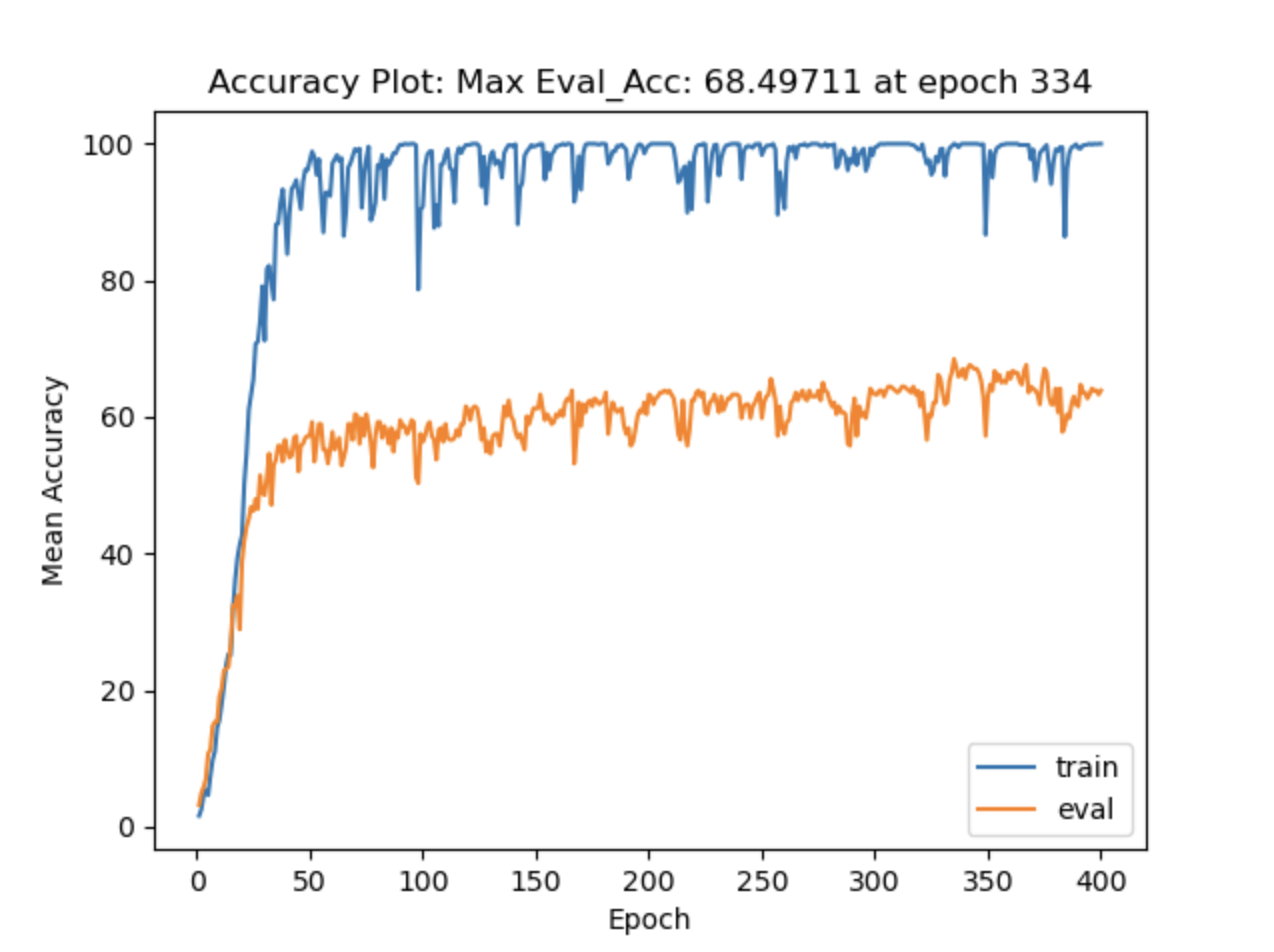}
        \label{fig:acc}
    \end{subfigure}
    \vspace{-0.5em}
    \caption{Top-1 accuracy of the final model on the validation set.}
    \label{fig:final_model_acc_both}
    \vspace{-1em}
\end{figure}

Using the best hyperparemeters found in the previous sections, we trained the final model on the downsampled dataset. The accuracy, top5 accuracy, and loss are displayed in Figure \ref{fig:final_model_acc_both} and Appendices \ref{top5_acc_final_model} and \ref{app_final_loss}. The training accuracy gets as high as 100\%. However, the validation accuracy tapers off at 68.50\%, underscoring the need for a larger training set to improve generalization. 

\section{Conclusion}
\label{sec:conclusion-future}
In this work, we introduce a novel approach for isolated sign language recognition (ISLR) by integrating video, segmentation, and pose estimation. Unlike prior methods that rely solely on RGB input or skeleton data, our architecture also leverages a segmentation pipeline to focus the model's attention on the most informative regions. This input is then processed by a deep neural network with elements to understand both the spatial and temporal dimensions.

Unfortunately, our available computing resources were insufficient for training the complete model on the full dataset. In response, we downsampled the dataset and temporarily removed the segmented video input. This enabled feasible experimentation but constrained our ability to fully test the generalization capacity of the model.

Despite these limitations, our approach achieved validation accuracy significantly higher than baselines and approached state-of-the-art results on the downsampled dataset, with a validation accuracy of 68.5\% and top-5 accuracy approaching 90\%. This demonstrates the model’s strong potential when trained with adequate resources. However, the persistent gap between training and validation performance underscores the need for larger training sets and further regularization.

Future work will scale training to the complete dataset, once additional GPU and memory resources are available. Additionally, revisiting the integration of visual features via ResNets and ViTs, exploring alternative temporal architecrtures, and grid searches on the hyperparameters will further enhance the model’s performance. Ultimately, this research lays the foundation for developing an application to provide real-time ASL translations, significantly bridging communication gaps for deaf signers.
%%----------------------------------------------------------------------------------%%
%% 								PAPER ACKNOWLEDGEMENT
%%----------------------------------------------------------------------------------%%
%\section*{Acknowledgement}
%The preferred spelling of the word ``acknowledgement'' in America is without 
%an ``e'' after the ``g''. Avoid the stilted expression ``one of us (R. B. 
%G.) thanks $\ldots$''. Instead, try ``R. B. G. thanks$\ldots$''. Put sponsor 
%acknowledgements in the unnumbered footnote on the first page.

%%----------------------------------------------------------------------------------%%
%% 							BIBLOGRAPHY (PAPER REFERENCES)
%%----------------------------------------------------------------------------------%%
\newpage
\bibliographystyle{IEEEtran}
\bibliography{IEEEabrv}

\appendices
\section{}
\label{app_original_model}

\begin{figure}[h]
    \begin{center}
        \includegraphics[width=1\linewidth]{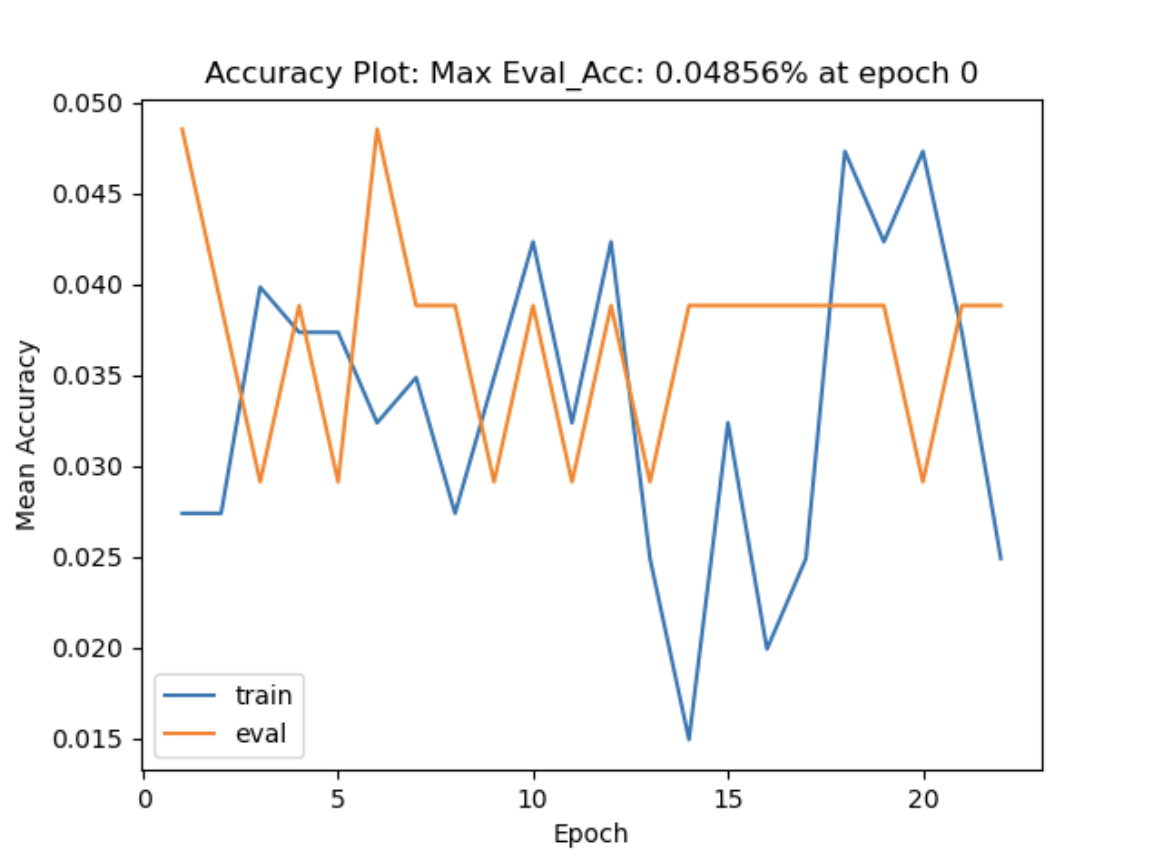}
        \caption{Accuracy of the original model through training. The model was only able to train for 24 epochs after two days of training. Validation accuracy never moved past the baseline of 0.05\%}\label{fig:original_results}
    \end{center}
    \vspace{-1em}
\end{figure}

The results for the original model (Section \ref{sec:proposed_model}) trained on the entire dataset are displayed in Figure \ref{fig:original_results}.

\section{}
\label{app_proposed_model}

\begin{figure}[h]
    \begin{center}
        \includegraphics[width=1\linewidth]{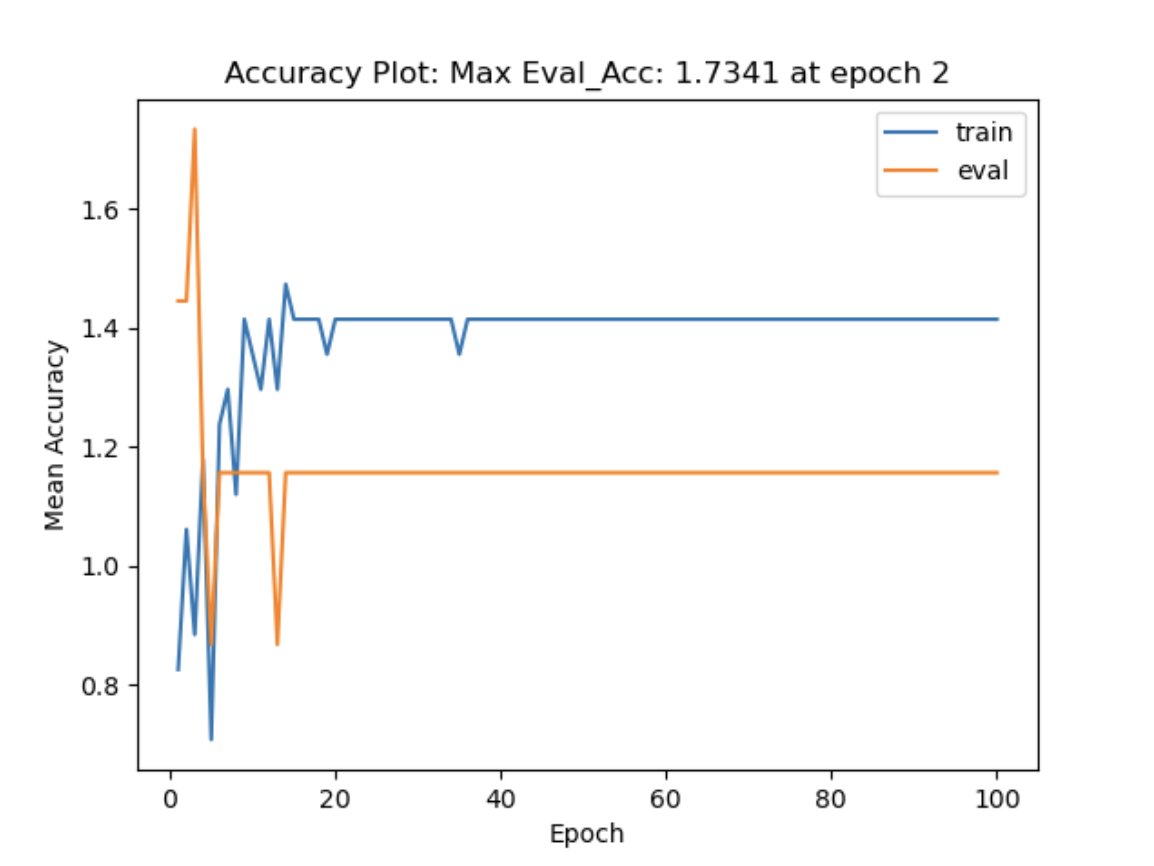}
        \caption{Accuracy of the model that uses both the segmented videos and the coordinates as input. The model passes the video frames through a ResNet 18, concatenates these features with the coordinates, and passes them through a transformer on the temporal dimension. The model did not improve its training or validation accuracy during training.}\label{fig:original_results_small}
    \end{center}
    \vspace{-1em}
\end{figure}

The results for the original model (Section \ref{sec:proposed_model}) trained on the downsampled dataset are displayed in Figure  \ref{fig:original_results_small}.

\section{}
\label{top5_acc_final_model}

\begin{figure}[h]
    \includegraphics[width=1\linewidth]{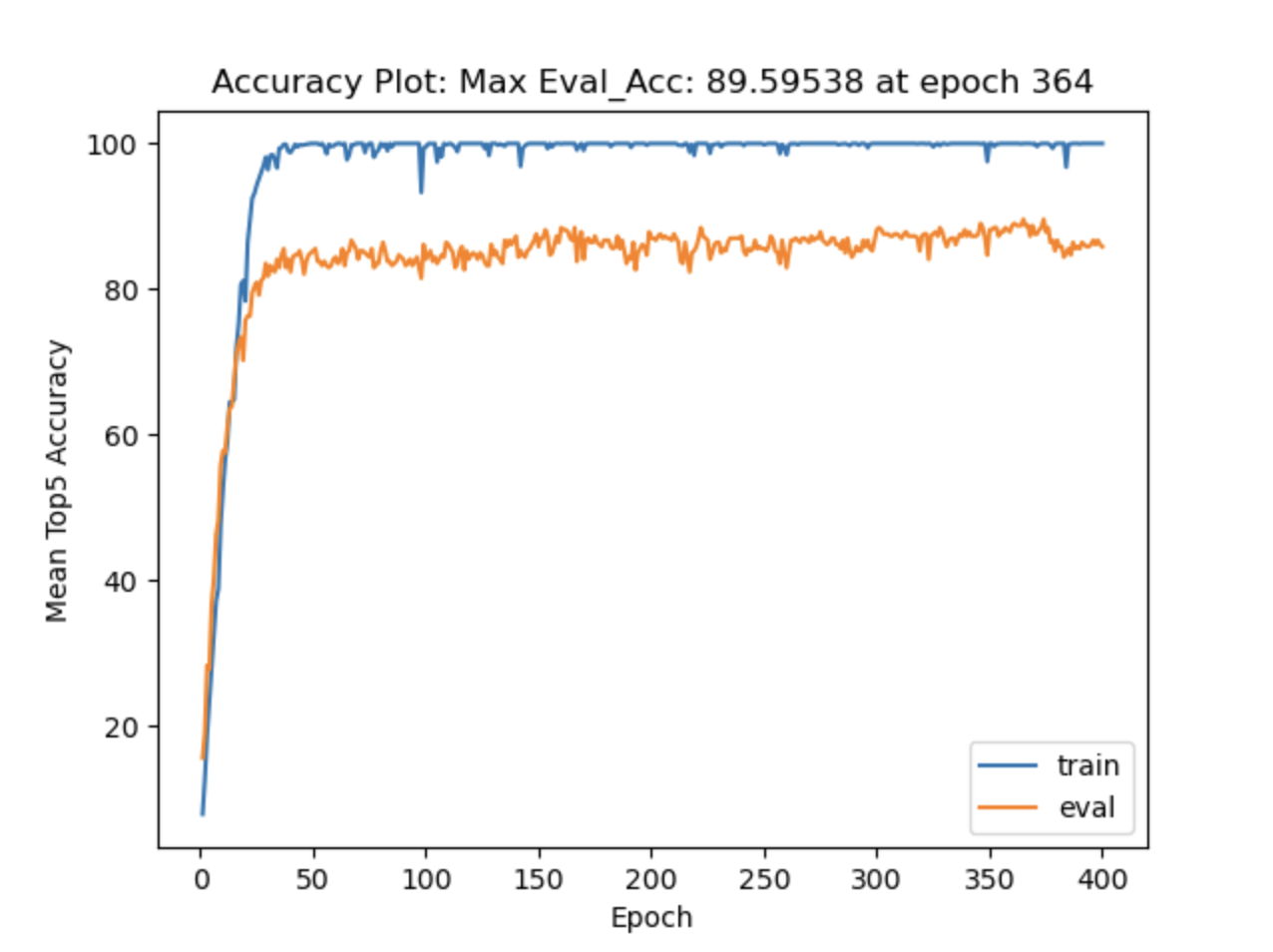}
    \caption{Top-5 accuracy of the final model on the validation set.}
    \label{fig:top5acc}
\end{figure}

Using the best hyperparemeters, we trained the final model on the downsampled dataset. The accuracy and top5 accuracy are displayed in Figures \ref{fig:final_model_acc_both} and \ref{fig:top5acc}.

\section{}
\label{app_final_loss}

\begin{figure}[h]
    \begin{center}
        \includegraphics[width=1\linewidth]{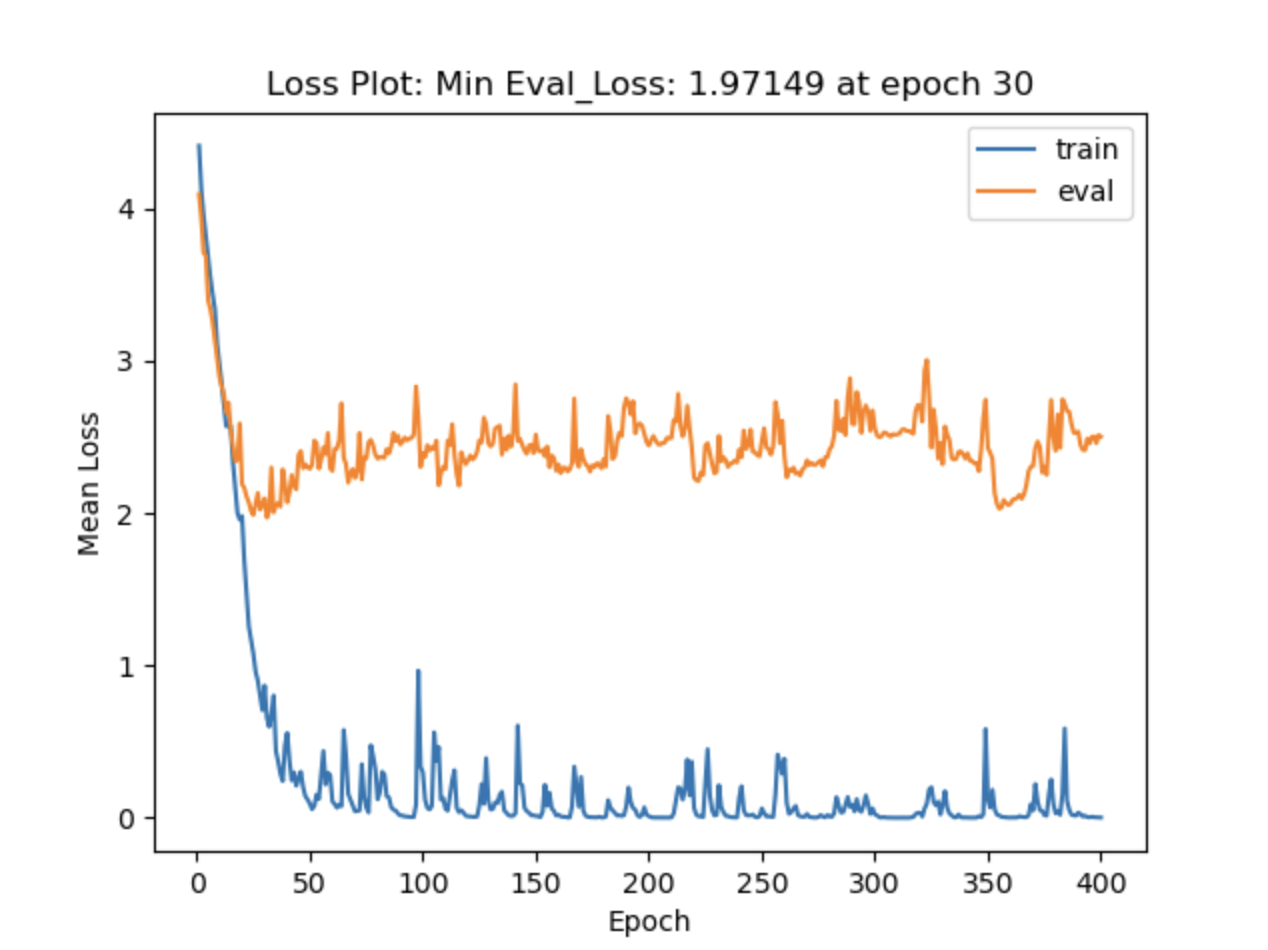}
        \caption{The training and validation loss of the model during training.}\label{fig:final_model_loss}
    \end{center}
    \vspace{-1em}
\end{figure}

The loss of the final model throughout training is shown in Figure \ref{fig:final_model_loss}. The training loss steadily decreases, until it bounces around $0$. However, the validation loss stops decreasing after about 30 epochs and remains much higher than the training loss.

\section{}
\label{the_dataset}

The \href{https://www.microsoft.com/en-us/research/project/asl-citizen/}{ASL Citizen dataset} (Section \ref{sec:the_dataset}) is a crowd sourced data set; it is sourced from everyday people in everyday conditions \cite{desai2023aslcitizen}. This helps the model generalize beyond controlled settings and perform effectively in everyday environments. The inclusion of diverse backgrounds further enhances robustness, enabling the model to operate reliably across a wide variety of conditions.  

\begin{figure}[h]
    \begin{center}
        \includegraphics[width=1\linewidth]{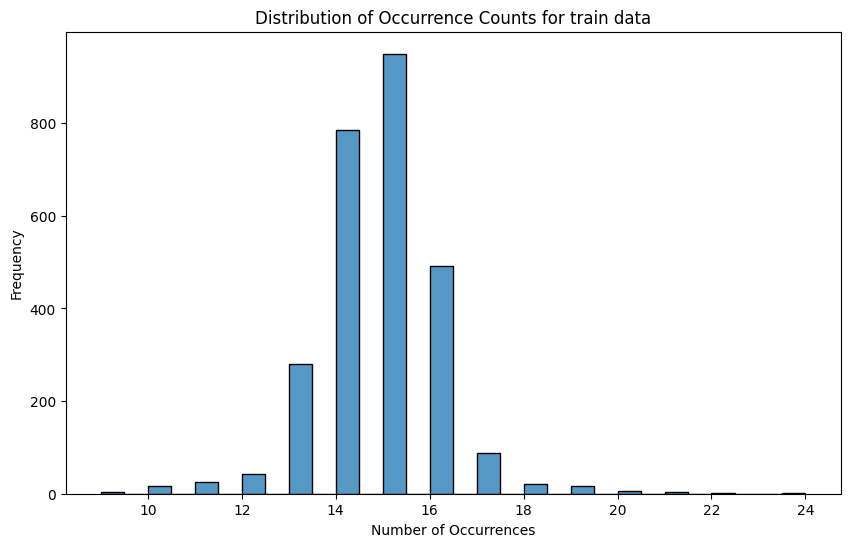}
        \caption{Distribution of word occurrences in the training dataset. Each vocabulary term appears between 9 and 24 times, with the majority occurring around 14 to 16 times.}\label{fig:distribution}
    \end{center}
    \vspace{-1em}
\end{figure}

The dataset is split into three parts: a training set, test set, and a validation set. Each set exhibits approximately Gaussian distributions for the number of occurences of each word, reflecting the natural variability of word usage in everyday language. This distribution helps the model generalize by emphasizing commonly used words over those that appear less frequently in daily communication. Figure \ref{fig:distribution} illustrates the distribution of the number of occurrences of each gloss in the training dataset. The mean number of instances of each gloss is 14.70, with a median of 15 and standard deviation of 1.31. The gloss which occurs the least is BEE2 which occurs 9 times. The gloss with the most occurrences is DOG1 with 24 instances. 

% For the appendix:
% The above is the distribution for the number of occurrences of each gloss in the validation dataset which has a total of 10304 videos. The mean number of instances of each gloss for the validation data is 3.77, with a standard deviation of 0.71. The median number of occurrences for each gloss is 4. The gloss which occurs the least is UNDERWEAR1 with 2 instances. The gloss with the most occurrences is ABOUT1 with 7 instances.

% The above is the distribution for the number of occurrences of each gloss in the testing dataset which has a total of 32941 videos. The mean number of instances of each gloss for the validation data is 12.06, with a standard deviation of 1.19. The median number of occurrences for each gloss is 12. The gloss which occurs the least is TYPE2 with 7 instances. The gloss with the most occurrences is BASKETBALL1 with 20 instances. 

% The majority of the videos are from people that are fluent in ASL with 34, 9, and 6 being at levels 7, 6 and 5 respectively. The data set also contains some people at the general proficiency with 1 person at each of levels 3 and 4.
%%----------------------------------------------------------------------------------%%
\end{document}